\documentclass[conference]{IEEEtran}

\usepackage[cmex10]{amsmath}
\usepackage{amsfonts}
\usepackage[ruled,vlined]{algorithm2e}

\newcommand\ceil[1]{\lceil#1\rceil}

\usepackage[noadjust]{cite}

\usepackage[hidelinks, bookmarks=false]{hyperref}
\usepackage{balance}
\usepackage{siunitx}
\usepackage{color}
\usepackage{booktabs} % For formal tables
\usepackage[T1]{fontenc} % fixing stupid way of showing "<" and ">"
\usepackage[]{graphicx}
\usepackage{multirow} % multirow in table
\renewcommand\baselinestretch{0.94}
\newcommand\CC{C\nolinebreak[4]\hspace{-.05em}\raisebox{.4ex}{\relsize{-3}{\textbf{++}}}}

\usepackage{cmap}
\usepackage{mmap}
\usepackage{microtype}
\DisableLigatures[f]{encoding=*,family=*}

\begin{document}
	\bstctlcite{IEEEexample:BSTcontrol}
	
	\title{\vspace{-0.3em}
		FFT-Based Deep Learning Deployment in Embedded Systems \vspace{-1.2 em}
	}
	
\author{
	\normalsize
	Sheng Lin$^{1*}$\thanks{Sheng Lin and Ning Liu contributed equally to this work.}, Ning Liu$^{1*}$, Mahdi Nazemi$^2$,  Hongjia Li$^1$, Caiwen Ding$^1$, Yanzhi Wang$^1$, and Massoud Pedram$^2$  \\
	\normalsize{$^1$Dept. of Electrical Engineering \& Computer Science, Syracuse University, Syracuse, NY, USA}\\
	\normalsize{$^2$Dept. of Electrical Engineering, University of Southern California, Los Angeles, CA, USA}\\   	
	\normalsize{ $^1$\{shlin,nliu03,hli42,cading,ywang393\}@syr.edu,$^2$\{mnazemi,pedram\}@usc.edu}
	}
	
	\maketitle
	\let\thefootnote\relax\footnote{*S. Lin and N. Liu contributed equally to this work.}
\begin{abstract}
Deep learning has delivered its powerfulness in many application domains, especially in image and speech recognition. As the backbone of deep learning, deep neural networks (DNNs) consist of multiple layers of various types with hundreds to thousands of neurons. Embedded platforms are now becoming essential for deep learning deployment due to their portability, versatility, and energy efficiency. The large model size of DNNs, while providing excellent accuracy, also burdens the embedded platforms with intensive computation and storage. Researchers have investigated on reducing DNN model size with negligible accuracy loss. This work proposes a Fast Fourier Transform (FFT)-based DNN training and inference model suitable for embedded platforms with reduced asymptotic complexity of both computation and storage, making our approach distinguished from existing approaches. We develop the training and inference algorithms based on FFT as the computing kernel and deploy the FFT-based inference model on embedded platforms achieving extraordinary processing speed.

%Deep neural networks with a large number of parameters perform powerfully in many different platforms. However, when it comes to embedded systems, the limit of memory storage and speed performance become a bottleneck for real-time applications. To demonstrate the effectiveness of our approach, we present Fast Fourier Transform-based deep neural networks to accelerate the image processing in embedded system. The new system architecture consists of fast training and inference methods. With Fast Fourier Transform (FFT) being the key computing kernel in the proposed approach, its recursive property is used as the key computation in our systems. It is by utilizing the structured circulant matrix, which significantly reduce the time complexity and memory storage. We show the new method to improve the performance of the image recognition on some benchmark data sets.  
\end{abstract}

\section{Introduction}
    
Recently deep learning has outstood from traditional machine learning techniques in many application areas, especially in image and speech recognition \cite{he2016deep,graves2013speech}.
The excellence of deep learning has also resulted in explorations of several emerging real-world applications, such as self-driving systems \cite{huval2015empirical}, automatic machine translations \cite{collobert2008unified}, drug discovery and toxicology \cite{burbidge2001drug}.
The deep learning is based on the structure of deep neural networks (DNNs), which consist of multiple layers of various types and hundreds to thousands of neurons in each layer.
Recent evidence has revealed that the network depth is of crucial importance to the success of deep learning, and many deep learning models for the challenging ImageNet dataset are sixteen to thirty layers deep~\cite{he2016deep}.
Deep learning achieves significant improvement in overall accuracy by extracting complex and high-level features at the cost of considerable up-scaling in the model size. 

In the big data era and driven by the development of semiconductor technology, embedded systems are now becoming an essential computing platform with ever-increasing functionalities.
At the same time, researchers around the world from both academia and industry have devoted significant efforts and resources to investigate, improve, and promote the applications of deep learning in embedded systems ~\cite{han2015deep}. 
Despite the advantages in DNN recognition accuracy, the deep layered structure and large model size of DNNs also increase computational complexity and memory requirement.
Researchers are faced with the following challenges when deploying deep learning models on embedded systems:
(i) Confined by the communication bandwidth of embedded systems, which are usually mobile terminals, it is still challenging to download large-size DNN models, even which can be offline-trained in data centers.
(ii) The large model size of deep learning also imposes stringent requirements on the computing resources and memory size of embedded systems.  

%As a result, it is impractical to implement the current deep neural networks on embedded systems due to several reasons as follows: (i) consider that a large portion of embedded systems are the mobile terminals, which have limited download bandwidth to receive the data. (ii) In this case, even the DNN models off-line are trained in data centers, the efficient and reliable downloads of those models are still greatly impeded by their large sizes. The most frequent computations in DNNs are the weight-involved operations such as inner products, and the inference of large neural networks results in extensive memory access and computational requirements, which in turn cause high energy consumption and strict requirement on memory size.
	
Motivated by these challenges, it is intuitive to implement a reduced-size deep learning model with negligible accuracy loss.
In fact, the state-of-the-art DNNs are often over-parameterized, hence the removal of redundant parameters in the deep learning models, if performed properly, will produce similar overall accuracy as the original models ~\cite{he2016deep}. 
Encouraged by this discovery, various deep learning model compression approaches have been investigated  ~\cite{han2015deep,ren2017sc,lecun1989optimal,pratt1989comparing,nazemi2017high}, including weight precision reduction, network pruning, weight matrix factorization, etc.
In this work, we propose a Fast Fourier Transform (FFT)-based DNN training and inference model suitable for embedded systems due to reduced asymptotic complexity of both computation and storage.
Our approach has obvious advantages over existing works on deep learning model compression e.g.,~\cite{han2015deep,lecun1989optimal,pratt1989comparing} in that those approaches result in an irregular network architecture that increases training and inference computation time, while our approach facilitates computation. Please also note the our proposed framework is distinct from the prior work of using FFT for convolutional layer acceleration by LeCun et al. ~\cite{mathieu2013fast}, because this prior work can only achieve convolutional layer acceleration instead of simultaneous compression.
We develop the training and inference algorithms based on FFT as the computing kernel and deploy the FFT-based inference model on embedded platforms. Experimental test results demonstrate that our model provides the optimization in different languages and achieve a significant improvement.

%they mainly focus on constructing an irregular network after applying weight storage reduction, which results in increased computation time, while we present specific algorithmic design optimization, mainly about weight storage, in order to achieve energy efficiency and high performance on embedded systems. 
%We develop the training and inference algorithms and perform various experiments on deep neural network's inference framework on embedded systems. Experimental results show that the model performs efficiently and gets reasonable accuracy.

\section{Related Work}
	
%     Over the last decade, large-scale deep neural networks (DNNs) have made
% breakthroughs in many fields, such as image recognition \cite{krizhevsky2012imagenet,lecun2015lenet,he2016deep}, speech recognition \cite{hinton2012deep,dahl2012context}, game playing, complicated control systems \cite{mnih2013playing,silver2016mastering,mnih2015human}, driver-less cars \cite{schmidhuber2015deep,huval2015empirical} and unmanned aerial systems (UAS) \cite{makantasis2015deep}. 
% DNNs have two promising applications:
% {\em 1)} Embedded and/or wearable devices for IoT, mobile and edge computing,
% which can potentially enable the future smart space and ubiquitous intelligence \cite{ma2005towards,dillon2009web};
% {\em 2)} Robotics, unmanned cars and UAS, which may become the 
% first step of the future fully autonomous systems and autonomous space \cite{tomic2012toward,sridharan2007color}. 
% Both applications require {\em efficient DNN implementations} that
% can overcome the challenges of high computational 
% complexity and large model 
% size~\cite{karpathy2015deep,catanzaro2013deep,simonyan2014very}.
    
Over the past decade, 
% many investigations on hardware accelerations in DNNs have been conducted~\cite{ren2017sc, chen2014diannao,esser2016convolutional}. 
a substantial number of techniques and strategies have been proposed to compress neural network size.  Weight pruning~\cite{han2015deep} is a well-known effective  approach, in which many weights with values of 0 are pruned to achieve high compression ratio. Other techniques such as threshold setting~\cite{han2015deep}, biased weight decay~\cite{pratt1989comparing}, etc., could be integrated to the weight pruning procedure.	     
% In order to deploy efficient neural networks, lots of different techniques and strategies are proposed to reduce the model size of neural networks. 
% The early work about Optimal Brain Damage was proposed in 1989, which uses the second derivative to delete part of weights during the training phase~\cite{lecun1989optimal}. 
% Considering the fact that a large portion of weights in a trained DNN model are close to zero, weight pruning is an efficient approach to achieve high compression ratio.
% Some other techniques, such as threshold setting~\cite{han2015deep}, biased weight decay~\cite{pratt1989comparing}, etc., can also be integrated to the pruning procedure for reduced model size.	
Another simple and popular approach to DNN model compression is the low-rank approximation of the weight matrix~\cite{denil2013predicting}. 
To overcome the potential high accuracy loss after low-rank approximation, ~\cite{chung2016simplifying} proposed to perform fine-tuning for the post-factorization of low-rank weight matrices to retain accuracy .
Lowering the presentation precision of weights is also an straightforward technique to reduce both the model size and computation cost of DNNs.
A fixed-point implementation was explored to replace the original floating-point models~\cite{anwar2015fixed}. 
Furthermore, designs with ultra-low precision weights, such as binary (-1 / +1) or ternary (-1 / 0 / +1) representation were proposed~\cite{courbariaux2016binarized,hwang2014fixed}.
By exploring the local and global characteristics of the weight matrix, weight clustering was proposed to reduce the number of weights linearly~\cite{gong2014compressing}. 
In addition, with the aid of gradients clustering in the training phase, the accuracy loss incurred by the weight clustering can be negligible~\cite{han2015deep}. 
	
%Distinguished from the above-mentioned works, this paper utilizes the structured matrices to construct DNN models. 
Some recent works adopted structured weight matrices in order to reduce the model size. 
In ~\cite{sindhwani2015structured}, weight matrices of fully-connected (FC) layers were constructed in the Toeplitz-like format to remove the redundancy of the DNN model. 
In \cite{cheng2015exploration}, the circulant matrix was introduced to enable further reduction in model size. 
An $n$-by-$n$ circulant matrix has a smaller number of parameters i.e., $n$ than that of a same-size Toeplitz matrix i.e., $2n$. 
In this work, we generalize the structured weight matrix method in that (1) we utilize block-circulant matrices for weight matrix representation, which achieves a trade-off between compression ratio and accuracy loss; (2) we extend the structured matrix method to convolutional (CONV) layers besides the FC layers; (3) we propose FFT-based DNN training and inference model and algorithm, which is highly suitable for  deployment in embedded systems; and (4) we implement and test the FFT-based DNN inference in various embedded platforms. 
%Motivated by these promising works, this paper investigates the advantage of significant reduction in computational operations and weight storage by applying the general block-circulant matrices, and optimizes this deep neural networks for embedded systems. 
     
\section{Background}
    
In this section, we introduce basic concepts of deep neural networks (DNNs), Fast Fourier Transform (FFT), and structured matrices, as the background of our proposed FFT-based training and inference algorithms. 
Specifically, we explain the various DNN layer types, the Cooley-Tukey algorithm for FFT, and the block-circulant matrices as the adopted structured matrices. 
    
\subsection{Deep Neural Networks}

Deep neural networks (DNNs) are distinguished from other types of neural networks by their depth and have dramatically improved the state-of-the-art in speech recognition, object detection, etc. 
Some commonly adopted DNN models include deep convolutional neural networks, deep belief networks, and recurrent neural networks. 
Despite the various network topologies targeting for different applications, these DNN models comprise of multiple functional layers with some commonly used structures. 
Following are the most commonly used layer structures in the state-of-the-art DNN models:

\emph{\textbf{The fully-connected (FC) layer}} is the most storage-intensive layer in DNN models \cite{qiu2016going} since each of its neurons is fully connected with all the neurons in the previous layer. 
The computation procedure of a FC layer consists of matrix-vector arithmetics (multiplication and addition) and transformation by the activation function, described as follows:
\begin{equation}
\bf{y}=\psi(\bf{W}^T\bf{x}+\mathbf{\theta})
\end{equation}
where $\bf{y}$ and $\bf{x}$ are outputs of this layer and the previous layer, respectively; $\textbf{W}\in \mathbb{R}^{m\times n}$ is the weight matrix of the synapses between this FC layer (with $n$ neurons) and its previous layer (with $m$ neurons); $\mathbf{\theta}\in \mathbb{R}^{n}$ is the bias vector; and $\psi(\cdot)$ is the activation function. The Rectified Linear Unit (ReLU) $\psi(x)=\max(0,x)$ is the most widely utilized activation function in DNNs.

\emph{\textbf{The convolutional (CONV) layer}}, as the name implies, performs two-dimensional convolution of its input to extract features that will be fed into subsequent layers for higher-level feature extracting. 
A CONV layer is associated with a set of learnable filters \cite{lecun1998gradient}, which are activated when specific types of features are found at some spatial positions from the inputs. 
Filter-sized moving windows are applied to the inputs to obtain a set of feature maps, by calculating the convolution of the filter and inputs in the moving window. 
Each \emph{convolutional neuron}, representing one pixel in a feature map, takes a set of inputs and the corresponding filter weights to calculate the inner-product. 
Given input feature map $\textbf{X}$ and the $r\times r$-sized filter (i.e., the \emph{convolutional kernel}) $\textbf{F}$, the output feature map $\textbf{Y}$ is calculated as
\begin{equation}
y_{a,b}=\sum_{i=1}^r\sum_{j=1}^r x_{a+i-1,b+j-1}\times f_{i,j},
\end{equation}
where $y_{a,b}$, $x_{a+i-1,b+j-1}$, and $f_{i,j}$ are elements in $\textbf{Y}$, $\textbf{X}$, and $\textbf{F}$, respectively. 
Multiple convolutional kernels can be adopted to extract different features in the same input feature map. 
Multiple input feature maps can be convolved with the same filter and results are summed up to derive a single feature map.
    
\subsection{Fast Fourier Transforms}
	
\begin{figure}[t]
		\centering
		\includegraphics[width = 0.45\columnwidth]{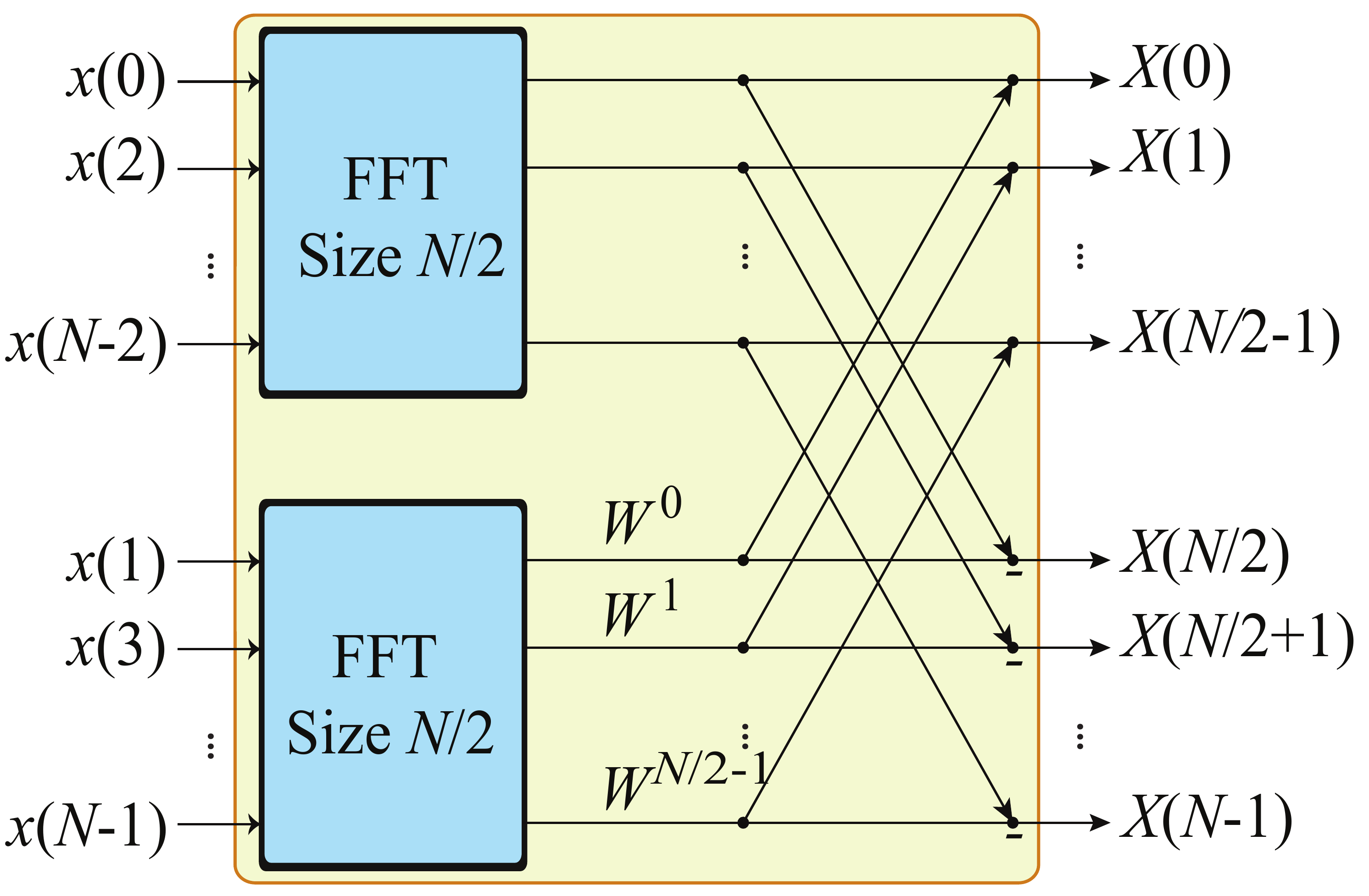}
		\caption{Illustration of Cooley-Tukey algorithm of FFT.}
		\label{fig:FFTRecursive}
	\end{figure}

The Fast Fourier Transform (FFT) is an efficient procedure for computing the discrete Fourier transform (DFT) of time series. 
It takes advantage of the fact that the calculation of the coefficients of the DFT can be carried out iteratively, which results in a considerable savings of computation time. 
The FFT not only reduces the computational complexity, but also substantially reduces round-off errors associated with these computations. 
In fact, both the computation time and round-off error are essentially reduced by a factor of $n/(log_{2} n)$ where $n$ is the number of data samples in the time series~\cite{cochran1967fast}.
Fig.~\ref{fig:FFTRecursive} shows the simplest and most common form of FFT, which is based on the Cooley-Tukey algorithm \cite{cooley1965algorithm}. 
It uses a divide and conquer approach to recursively break down the DFT of an arbitrary composite size $N = N_1\cdot N_2$ into many smaller DFTs of sizes $N_1$ and $N_2$, in order to reduce the computation time to O($n\log n$) for highly composite $N$~\cite{cooley1965algorithm}.
	
\subsection{Structured Matrices}
	
An $n$-by-$m$ matrix $\mathbf{A}$ is called a structured matrix when it has a low displacement rank $\upsilon$~\cite{sindhwani2015structured}. One of the most important characteristics of structured matrices is their low number of independent variables. The number of independent parameters is $O(n)$ for an $n$-by-$n$ structured matrix instead of $O(n^2)$, which indicates that the storage complexity can be potentially reduced to $O(n)$. As a representative example, a circulant matrix $\mathbf{W}\in\mathbb{R}^{n \times n}$ is defined by a vector $\mathbf{w}=(w_1,w_2,...,w_n)$ as follows:\\
	\[
	\begin{bmatrix}
	w_{1} & w_{n} & \dots & w_{3} & w_{2} \\
	w_{2} & w_{1} & w_n & & w_{3} \\
	\vdots & \vdots & \vdots & \ddots & \vdots \\
	w_{n-1} &  & \ddots & \ddots  & w_{n} \\
	w_{n} & w_{n-1} & \dots  & w_{2} & w_{1}
	\end{bmatrix}
	\]
% More precisely, with the proper choice of operator matrices $\mathbf{M}$ and $\mathbf{N}$, if the Sylvester displacement $\nabla_{\mathbf{M},\mathbf{N}}(\mathbf{A}):=\mathbf{MA}-\mathbf{AN}$ %\textcolor{red}{(where is N? Sheng: Please see the reference~\cite{sindhwani2015structured}. (1) Are you citing the right reference? I don't find it there. (2) When you search Sylvester displacement, it is not like that. Sheng: Wrong citation, and update the equation.)} 
% and the Stein displacement $\nabla_{\mathbf{M},\mathbf{N}}(\mathbf{A}):=\mathbf{A}-\mathbf{MAN}$ of matrix $\mathbf{A}$ have a rank $\upsilon$ bounded by a value that is independent of the size of $\mathbf{A}$, then matrix $\mathbf{A}$ is referred to as a structured matrix with a low displacement rank ~\cite{sindhwani2015structured}. This statement is true even when matrix $\mathbf{A}$ itself is a full-rank matrix. There are a series of commonly used structured matrices, including circulant matrix, Toeplitz matrix, Vandemonde matrix, etc. 

The definition and analysis of structured matrices have been generalized to the case of $m$-by-$n$ matrices where $m \neq n$, e.g., the block-circulant matrices.
Besides, the computational complexity for many matrix operations, such as matrix-vector multiplication, matrix inversion, etc., can be significantly reduced when operating on structured matrices. 

\section{Fast Fourier Transform-Based DNN Model}
    
In this section, we propose an efficient inference algorithm and explain the training algorithm in deep neural networks by using block-circulant matrices. We achieve a simultaneous and significant reduction in computational complexity of inference and training processes, and also weight storage. Besides, we have performed theoretical analysis to prove the effectiveness of substituting matrix multiplication with the Fast Fourier Transform method and utilizing block-circulant matrices, thereby guaranteeing applicability of the proposed framework on a wide variety of applications and emerging deep learning models.
    
\subsection{Block-Circulant Matrix-Based Inference and Training Algorithms for FC Layers}

% The fully-connected (FC) layer is the most storage-intensive layer in DNNs since each of its neurons is fully connected with neurons in the previous layer.
% The computation procedure of a FC layer consists of matrix-vector arithmetic and transformation by the activation function.

Cheng et al. proposed circulant matrix-based DNN training and inference algorithms for FC layers \cite{cheng2015exploration}.
However, in many practical applications such schemes cannot be directly used because:
(1) It is very common that the weight matrices of DNNs are non-square matrices due to the specific need of different applications; and (2) Even if the weight matrices are square, in many cases the compression is too aggressive and hence causes non-negligible accuracy loss.
To address the above challenges, we present the block-circulant matrix-based inference and training algorithms.
	
\begin{figure}[b]
		\centering
        \vspace{-0.7cm}
		\includegraphics[width = 0.55\columnwidth]{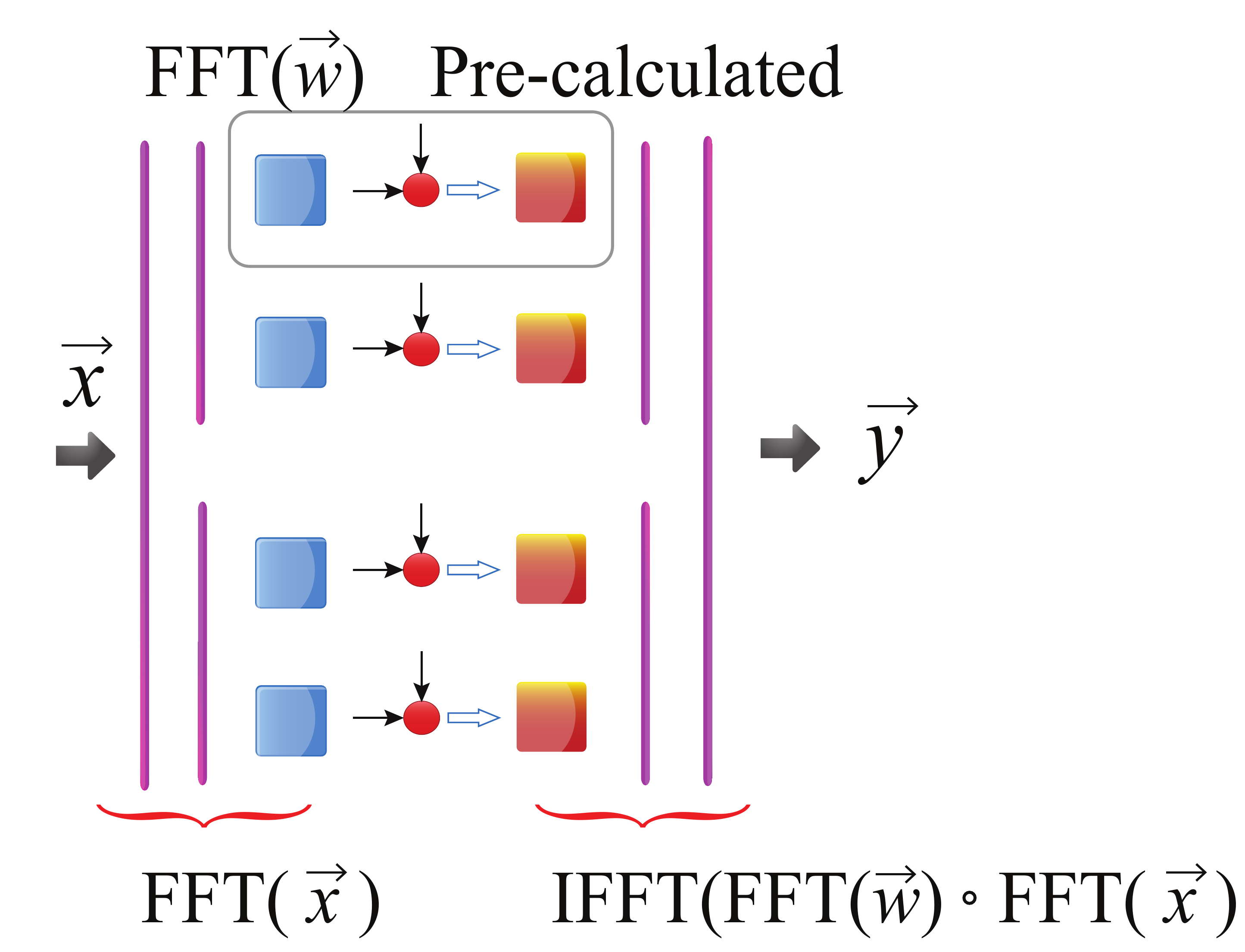}
		\caption{The ``FFT $\rightarrow$ component-wise multiplication $\rightarrow$ IFFT" procedure.}
		\label{fig:FFT}
	\end{figure}

Recall that the forward propagation during the inference phase of a FC layer is performed as
$\bf{y}=\psi(\bf{W}^T\bf{x}+\mathbf{\theta})$,
%$\mathbf{y}=f(\mathbf{a})$, 
where $\psi$ is the activation function, $\bf{W}$ is the weight matrix, $\bf{x}$ is the input vector, and $\mathbf{\theta}$ is the biases. 
The computation bottleneck is the calculation of $\bf{W}^T\bf{x}$. 
When using a block-circulant matrix for representing $\bf{W}$, a fast multiplication algorithm for $\bf{W}^T\bf{x}$ exists, which will result in a significant reduction in computational complexity.
Assume that the weight matrix is an $m$-by-$n$ block-circulant matrix $\mathbf{W}=[\mathbf{C}_1|\mathbf{C}_2|...|\mathbf{C}_k]^{\mathbf{T}}$; the input vector is $\mathbf{x}=(\mathbf{x}_1|\mathbf{x}_2| ...| \mathbf{x}_k)$; and the bias vector is $\mathbf{\theta}=(\mathbf{\theta}_1|\mathbf{\theta}_2|...|\mathbf{\theta}_k)$. Each circulant matrix $\mathbf{C}_i \in \mathbb{R}^{n\times n}$ is defined by a length-$n$ vector $\mathbf{w}_i=(w_{i,1},w_{i,2},...,w_{i,n})^{\mathbf{T}}$, $i\in\{1,...,k\}$, $m=kn$\footnote{For general values of $m$ and $n$, we can apply zero padding such that the definition of block-circulant matrices can be applied.}, and $\mathbf{x}_i=(x_{i,1},x_{i,2},...,x_{i,n})^{\mathbf{T}}$. Hence, $\bf{W}^T\bf{x}$, as the key computation bottleneck in the inference phase, can be simplified as below:
	\begin{equation}
		\mathbf{W}^{\mathbf{T}}\mathbf{x}=\sum_{i=1}^k \mathbf{C}_i\mathbf{x}_i=\sum_{i=1}^k \text{IFFT}\big(\text{FFT}(\mathbf{w}_i)\circ \text{FFT}(\mathbf{x}_i)\big)
	\end{equation}
	where $\text{FFT}$, $\text{IFFT}$, and $\circ$ represent a Fast Fourier transform (FFT), an inverse FFT, and an element wise multiplication, respectively. This ``FFT $\rightarrow$ component-wise multiplication $\rightarrow$ IFFT" procedure to implement $\bf{W}^T\bf{x}$ shown in Fig.~\ref{fig:FFT} is derived from the circular convolution theorem~\cite{pan2012structured,zhao2017theoretical}. The overall computational complexity in this FC layer will be O($n\log n$), achieving a significant reduction compared to O($n^2$) when calculating $\bf{W}^T\bf{x}$ directly. In order to store the weights for the inference phase, we can simply keep the FFT result $\text{FFT}(\mathbf{w}_i)$ (which is a vector) instead of the whole matrix $\mathbf{W}$, thereby reducing the storage complexity to O($n$) for an FC layer. Algorithm~\ref{alg:inference} summarizes the FFT-based inference algorithm.
	
	\SetAlgoNoLine
	\begin{algorithm}[]
		\label{alg:inference}
		\KwIn{$w, x, m, n$}
		\KwOut{$a$}
		$sa \gets max(m, n)$\;
		$si \gets min(m, n)$\;
		$k \gets \ceil{sa/si}$\;
		partition $w$ into $k$ vectors, $w_1, \dots, w_k$\;
		\eIf{$m$ > $n$} {
			\For{$i \gets 0$ until $k$}{
				$a$ $\gets $ $a$ + $\text{ifft}( \text{fft}(\mathbf{w_i}) \circ \text{fft}(x))$\;
			}
		} {
		partition $x$ into $k$ vectors, $x_1, \dots, x_k$\;
		\For{$i \gets 0$ until k}{
			$a$ $\gets $ $a$ + $\text{ifft}( \text{fft}(\mathbf{w_i}) \circ \text{fft}(x_i))$\;
		}
	}
	\Return{$a$}\;
	\caption{Block-circulant Matrix-based Inference}
	\end{algorithm}
    
	Besides the inference procedure, the reformulated training (weight updating) algorithm in the scenario of using block-circulant matrices will also result in significant accelerations. We denote $\mathbf{a}=\mathbf{W}^{\mathbf{T}}\mathbf{x}+\mathbf{\theta}=(\mathbf{a}_1|\mathbf{a}_2|...|\mathbf{a}_k)^{\mathbf{T}}$ and $\mathbf{a}_i=(a_{i,1},a_{i,2},...,a_{i,n})^{\mathbf{T}}$, then the weight updating rule for the block-circulant FC layer is given by:
	\begin{equation}
		\mathbf{w}_i\leftarrow \mathbf{w}_i-\epsilon\cdot \text{IFFT}\Big(\text{FFT}\big(\frac{\partial L}{\partial \mathbf{a}_i}\big)\circ \text{FFT}(\mathbf{x}'_i)\Big)\cdot \mathbf{I}
	\end{equation}
	where $J$, $\mathbf{I}$, $\epsilon$, and $\mathbf{x}'_i$ represent the loss function, an all-one column vector, the learning rate, and the base vector that defines the circulant matrix $\frac{\partial \mathbf{a}_i}{\partial \mathbf{w}_i}$ (which is formally derived), respectively. Notice that since $\frac{\partial \mathbf{a}_i}{\partial \mathbf{w}_i}$ is a circulant matrix, similar to inference, we can utilize the ``FFT$\rightarrow$component-wise multiplication$\rightarrow$IFFT" procedure to accelerate the matrix-vector multiplication. The computational complexity will be O($n\log n$) in each updating step in this layer, which is a significant reduction from O($n^2$) in traditional backpropagation procedure. Algorithm~\ref{alg:training} summarizes the FFT-based training algorithm.
	
	\SetAlgoNoLine
	\begin{algorithm}
		\label{alg:training}
		\KwIn{$\frac{\partial L}{\partial a}$, $w, x, m, n$}
		\KwOut{$\frac{\partial L}{\partial w}$, $\frac{\partial L}{\partial x}$}
		$sa \gets max(m, n)$\;
		$si \gets min(m, n)$\;
		$k \gets \ceil{sa/si}$\;
		partition $w$ into $k$ vectors, $w_1, \dots, w_k$\;
		partition $\frac{\partial L}{\partial w}$ into $k$ vectors, $\frac{\partial L}{\partial w_1}, \dots, \frac{\partial L}{\partial w_k}$\;
		\eIf{$m > n$} {
			partition $\frac{\partial L}{\partial a}$ into $k$ vectors, $\frac{\partial L}{\partial a_1}, \dots, \frac{\partial L}{\partial a_k}$\;
			\For{$i \gets 0$ until k}{
				$\frac{\partial L}{\partial w_i} \gets \text{ifft}(
				\text{fft}(\frac{\partial L}{\partial a})
				\circ 
				\text{fft}(\mathbf{x'}) 
				)
				\cdot 
				\mathbf{1}$\;
				$\frac{\partial L}{\partial x} \gets \frac{\partial L}{\partial x} + \text{ifft}( \text{fft}(\frac{\partial L}{\partial a})
				\circ
				\text{fft}(\mathbf{w_i'}) )$\;
			}
		} {
		partition x into $k$ vectors, $x_1, \dots, x_k$\;
		partition $\frac{\partial L}{\partial x}$ into $k$ vectors, $\frac{\partial L}{\partial x_1}, \dots, \frac{\partial L}{\partial x_k}$\;
		\For{$i \gets 0$ until k}{
			$\frac{\partial L}{\partial w_i} \gets \text{ifft}(
			\text{fft}(\frac{\partial L}{\partial a})
			\circ 
			\text{fft}(\mathbf{x_i'}) 
			)
			\cdot 
			\mathbf{1}$\;
			
			$\frac{\partial L}{\partial x_i} \gets \text{ifft}( \text{fft}(\frac{\partial L}{\partial a})
			\circ
			\text{fft}(\mathbf{w_i'}) )$\;
		}
	}
	\Return{$\frac{\partial L}{\partial w}$, $\frac{\partial L}{\partial x}$}\;
	\caption{Block-circulant Matrix-based Training}
	\end{algorithm}
    
	\subsection{Block-Circulant Matrix-Based Inference and Training Algorithms for CONV Layer}
    
	The use of block-circulant matrices can also enable significant reduction in computational and storage complexities of the Convolutional layer. The Convolutional layers are often associated with multiple input and output feature maps in DNNs. Therefore, the computation of the Convolutional layer is described using tensor format as follows:
\begin{equation}
\mathcal{Y}(x,y,p)=\sum_{i=1}^r\sum_{j=1}^r\sum_{c=1}^C\mathcal{F}(i,j,c,p)\mathcal{X}(x+i-1,y+j-1,c),
\end{equation}
where $\mathcal{X}\in\mathbb{R}^{W\times H\times C}$, $\mathcal{Y}\in\mathbb{R}^{(W-r+1)\times(H-r+1)\times P}$, $\mathcal{F}\in\mathbb{R}^{r\times r\times C\times P}$ denote the input, output, and weight ``tensors" of the Convolutional layer, correspondingly. $C$ is the number of input maps. $W$ and $H$ are the spatial dimensions of the input maps. $P$ is the total number of output maps, and $r$ is the size of the convolutional kernel.

We generalize the ``block-circulant structure" as rank-4 tensor ($\mathcal{F}$) in the Convolutional layer, i.e., \emph{each slice $\mathcal{F}(\cdot,\cdot,i,j)$ is a circulant matrix.} Then, we reformulate the inference and training algorithms of the Convolutional layer to matrix-based operations. 

\begin{figure}[htbp]
\begin{center}
\includegraphics[width = 0.33\textwidth]{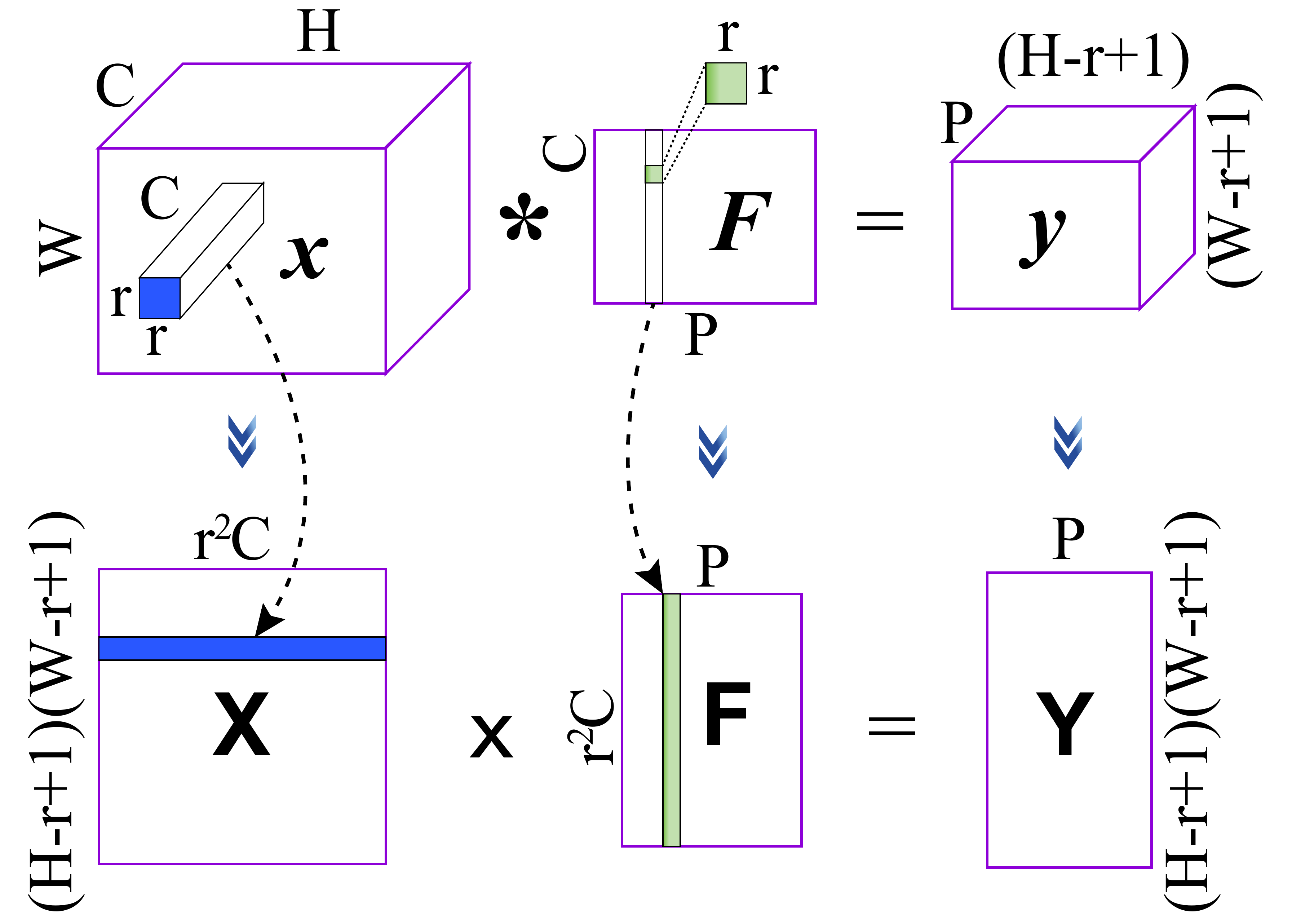}
\caption{Reformulation of Eqn. (3) to matrix multiplication.}
\label{fig_reformulation}
\end{center}
\vspace{-0.3em}
\end{figure}

In the Convolutional layer, to enhance the implementation efficiency, software tools provide an efficient approach of changing tensor-based operations to matrix-based operations equivalently \cite{jia2014caffe,vedaldi2015matconvnet}. Fig.~\ref{fig_reformulation} demonstrates the application of the method to reformulate Eqn. (3) to the matrix multiplication $\mathbf{Y}=\mathbf{XF}$, where $\mathbf{X}\in\mathbb{R}^{(W-r+1)(H-r+1)\times Cr^2}$, $\mathbf{Y}\in\mathbb{R}^{(W-r+1)(H-r+1)\times P}$, and $\mathbf{F}\in\mathbb{R}^{Cr^2\times P}$.

 Based on the reshaping principle between $\mathcal{F}$ and $\mathbf{F}$, we have:
\begin{equation}
f_{a+C(i-1)+Cr(j-1),b}=f_{C(i-1)+Cr(j-1),b-a}, \forall a, b
\end{equation}
where $\mathbf{F}$ is a block-circulant matrix. Therefore, the ``FFT$\rightarrow$component-wise multiplication $\rightarrow$IFFT" procedure can be applied to accelerate $\mathbf{Y}=\mathbf{XF}$, leading to the acceleration of (3). With the assist of the proposed approach, the computational complexity for (3) is reduced from O($WHr^2CP$) to O($WHQ\log Q$), where $Q=\max(r^2C,P)$.
    
	\section{Software Implementation}
	
   \begin{table*}[t]
		\centering
		\caption{Platforms under test and their specifications.}
		\begin{tabular*}{\textwidth}{l @{\extracolsep{\fill}} *{6}{c}}
			Platform        & Android          & Primary CPU                                & Companion CPU                              & CPU Architecture   & GPU  & RAM (GB) \\
			\midrule
			LG Nexus 5      & 6 (Marshmallow)  & 4 $\times$ 2.3\si{\giga\hertz} Krait 400   & -                                          & ARMv7-A            & Adreno 330 & 2 \\
			Odroid XU3      & 7 (Nougat)       & 4 $\times$ 2.1\si{\giga\hertz} Cortex-A15  & 4 $\times$ 1.5\si{\giga\hertz} Cortex-A7   & ARMv7-A            & Mali T628       & 2 \\
            Huawei Honor 6X & 7 (Nougat)  & 4 $\times$ 2.1\si{\giga\hertz} Cortex-A53  & 4 $\times$ 1.7\si{\giga\hertz} Cortex-A53  & ARMv8-A            & Mali T830       & 3
		\end{tabular*}
		\label{table:platforms}
	\end{table*}

	In this section, we provide detailed explanation of our software implementation, experimental setup, and evaluation of the proposed inference framework on various Android-based platforms with embedded processors and various datasets. The purpose of this software implementation is to reveal the potential of embedded systems in running real time applications that involve deep neural networks.
	
\begin{figure}[tb]
	\centering
	\includegraphics[width = 0.85\columnwidth]{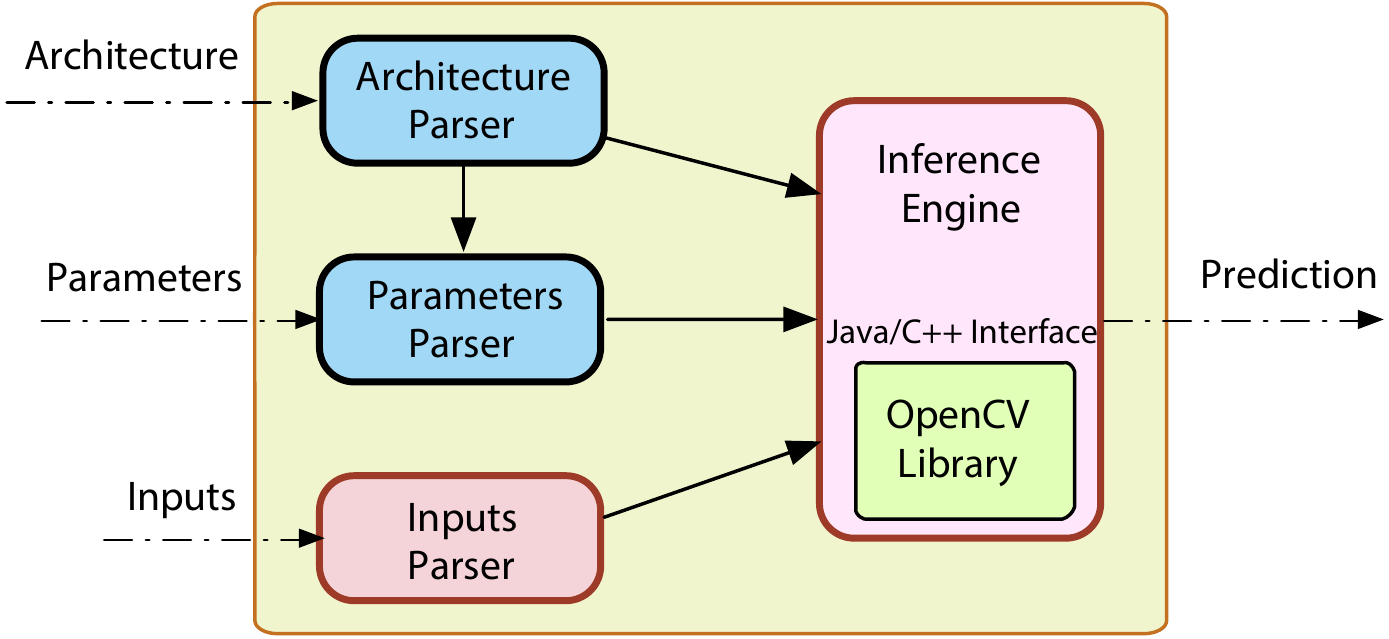}
	\caption{Building blocks of software implementation.}
	\label{fig:software_modules}
\end{figure}

	The software implementation of proposed inference framework for Android-based platforms is comprised of four high-level modules. The first module is responsible for constructing the network architecture. The second module reads a file that contains trained weights and biases. The third module loads test data that consists of input features and predefined classification labels, and finally, the fourth module performs inference for predicting labels. Fig.~\ref{fig:software_modules} depicts these high-level building blocks of the software implementation, along with their interactions. It should be noted that the test data may be loaded from a file, camera, etc.
    
% 	The basic operations required for implementing the inference module are matrix multiplication for fully-connected layers and convolutional layers, FFT and IFFT for layers with block-circulant weight matrices, element-wise exponentiation and division for a softmax layer, and various kinds of activation functions. 
    
    We utilize the OpenCV\cite{opencv_library} as core computing library in our project. OpenCV is an open-source cross-platform library of programming functions that is mainly targeted for computer vision applications and includes efficient implementation of aforementioned operations. OpenCV is written in \CC, and it provides the API (Application Program Interface) for both \CC and Java. We implement two versions of software for inference: one that uses OpenCV's Java API, which is more convenient for Android development, and another one that is developed in \CC~using Android NDK (Native Development Kit), uses OpenCV's \CC~API, and is expected to have a better performance.

	\subsection{Experimental Setup}
	
	We run the inference application on various platforms of different generations in order to evaluate the applicability of the inference on embedded systems. Table \ref{table:platforms} summarizes the specifications of test platforms.

	The OpenCV Manager is installed on all target platforms in order to link OpenCV libraries dynamically and reduce memory usage. Additionally, hardware specific optimizations are applied by OpenCV Manager for an application's supported platforms.
	
	In order to standardize the evaluation process on all platforms, the airplane mode is switched on to eliminate telecommunication overhead; all other running applications are closed to ensure they do not affect runtime; and the device is plugged in to avoid performance throttling applied by a platform's governor. Though this is the standard setup, we will study the performance of inference process in situations where the device is running on its battery.
    
	\subsection{MNIST}
    
	MNIST dataset \cite{mnist} is a handwritten digits dataset which includes 28$\times$28 greyscale images with 60,000 images for training and 10,000 images for testing. The original images in the MNIST dataset are resized using a bilinear transformation, and such transformation is used for both training and testing. Various neural network architectures are explored for each dataset and a few of them are presented in this paper.
    
	For the MNIST dataset, two different neural network architectures are evaluated. In the first architecture (Arch.~1), the input layer consists of 256 neurons that represent the resized MNIST images. The next two layers comprise of 128 neurons each and are based on block-circulant matrix based FC layers. Finally, the last layer is a softmax layer that consists of 10 neurons representing the ten possible predictions for the digits. The second architecture (Arch.~2) has 121 neurons in the input layer, 64 neurons in the two hidden layers, and similar to Arch.~1, a softmax layer as the output layer. Table~\ref{table:mnist_results} summarizes the runtime of each round of inference process using these architectures and on various mobile platforms.
	
% 	\begin{table}[h]
% 		\centering
% 		\caption{Core Runtime of each round of inference for resized MNIST images.}
% 		\label{table:mnist_results}
% 		\begin{tabular}{cccccc}
%         	\toprule
%         	\multirow{2}{*}{Architecture}     & \multirow{2}{*}{Implementation}     & \multirow{2}{*}{Accuracy (\%)}     &\multicolumn{3}{c}{Runtime (\si{\us} per image)}\\
% 			\cmidrule{4-6}
%             {}                                & {}                                   & {}                                      & XU3     & Honor 6X      \\
%             \midrule
% 			\multirow{2}{*}{Arch.~1}          & Java                                 & 95.47                                     & 294.1    &256.7  \\
% 			\cmidrule{2-6}
%             {}                                & \CC                                  & 95.47                                     & 122.0    &101.0    \\
% 			\midrule
% 			\multirow{2}{*}{Arch.~2}          & Java                                 & 93.59                                   & 278.2      &221.7  \\
% 			\cmidrule{2-6}
%             {}                                & \CC                                  & 93.59                                     & 119.1      &98.5  \\ 
%             \midrule
% 			{Baseline}          & \CC                               & 98.12                                & 167.8     &178.6  \\
%             \bottomrule
% 		\end{tabular}
% 	\end{table}
    
    \begin{table}[h]
		\centering
		\caption{Core Runtime of each round of inference for resized MNIST images.}
		\label{table:mnist_results}
		\begin{tabular}{cccccc}
        	\toprule
        	\multirow{2}{*}{Architecture}     & \multirow{2}{*}{Implementation}     & \multirow{2}{*}{Accuracy (\%)}     &\multicolumn{3}{c}{Runtime (\si{\us} per image)}\\
			\cmidrule{4-6}
            {}                                & {}                                   & {}                                 & Nexus 5     & XU3     & Honor 6X      \\
            \midrule
			\multirow{2}{*}{Arch.~1}          & Java                                 & 95.47                              & 359.6       & 294.1    &256.7  \\
			\cmidrule{2-6}
            {}                                & \CC                                  & 95.47                              & 140.0       & 122.0    &101.0    \\
			\midrule
			\multirow{2}{*}{Arch.~2}          & Java                                 & 93.59                              & 350.9       & 278.2      &221.7  \\
			\cmidrule{2-6}
            {}                                & \CC                                  & 93.59                              & 128.5       & 119.1      &98.5  \\ 
%             \midrule
% 			{Baseline}          & \CC                               & 98.12                              & 115.9       & 167.8     &178.6  \\
            \bottomrule
		\end{tabular}
	\end{table}
    Based on the results summarized in Table~\ref{table:mnist_results}, the \CC~implementation is about 60-65\% faster than the Java implementation. One of the reasons for this superior performance is related to memory limitations and management policy in Android. While applications written in \CC~have an unlimited heap size, Java applications are restricted to platform-specific heap sizes. As a result, a constraint is imposed on the amount of data that an application written in Java can deal with at each instance of time. 
%Additionally, Java's garbage collector needs to run from time to time in order to free up memory. This leads to a runtime overhead that a \CC~application does not suffer from.
    
    Another potential reason that may explain the considerable performance difference between the two implementations is the overhead due to switching from Java to \CC~and vice versa. Because the OpenCV library is written in \CC, it needs to covert data from \CC~data types to Java data types whenever the Java API is used. We believe that these conversions do not affect the runtime significantly, but can cause certain difference in performance across the two implementations.
    
    Considering different architectures mentioned in Table~\ref{table:mnist_results}, one can observe that going from the smaller network to a bigger network increases the accuracy by about 2\% while it increases the memory required for storing parameters by a factor of about two and increases the runtime of Java and \CC~implementations by about 2\% and 9\%, respectively.
    It should be noted that when the device is running on its battery, the runtime will increase by about 14\% in the Java implementation, but remains unchanged in the \CC~implementation.
    
%     Comparing the best result in our FFT based model with the baseline, it can see that our model significantly perform better than the baseline. Our model run 35-75$\%$ faster than the baseline with a negligible reduced accuracy \textcolor{red}{shall we remove the baseline and only compare with TrueNorth?}.

\subsection{CIFAR-10}
    The CIFAR-10 \cite{cifar10} dataset contains 32$\times$32 color images from 10 classes, where there are 50,000 training images and 10,000 testing images. The structure of deep neural network can be denoted as 128x3x32x32-64Conv3-64Conv3-128Conv3-128Conv3-512F-1024F-1024F-10F (Arch.~3). Here 128x3x32x32 represents that (i) the batch size is 128; (ii) the number of input channel is 3, (iii) and the feature size of input data is 32x32. In addition, 128Conv3 indicates that 128 3x3 convolutional filters are used in the convolutional layer. In addition, 512F or 10F means that the number of neurons in the FC layer is 512 or 10, respectively. In addition, both the original and compressed models are trained with learning rate 0.001 and momentum 0.9. In this network architecture, the first two convolutional layers are traditional convolutional layers (no block circulant, which is treated as preprocessing similar to the IBM TrueNorth paper~\cite{esser2016convolutional}). Based on the results summarized in Table~\ref{table:cifar10_results}, the \CC~implementation is about 130\% faster than the Java implementation. 

	\begin{table}[h]
		\centering
		\caption{Core Runtime of each round of inference process for cifar-10 images.}
		\label{table:cifar10_results}
		\begin{tabular}{cccccc}
        	\toprule
        	\multirow{2}{*}{Architecture}     & \multirow{2}{*}{Implementation}     & \multirow{2}{*}{Accuracy (\%)}     &\multicolumn{3}{c}{Runtime (\si{\us} per image)}\\
			\cmidrule{4-6}
            {}                                & {}                                   & {}                                  & XU3     & Honor 6X      \\
            \midrule
			\multirow{2}{*}{Arch.~3}          & Java                                 & 80.2                                   & 21032      &19785  \\
			\cmidrule{2-6}
            {}                                & \CC                                  & 80.2                                    & 8912      &8244  \\ 
            \bottomrule
		\end{tabular}
	\end{table}
    
    \subsection{Comparison Results on Performance and Accuracy}

\begin{figure}[t]
\begin{center}
\includegraphics[width = 0.3\textwidth]{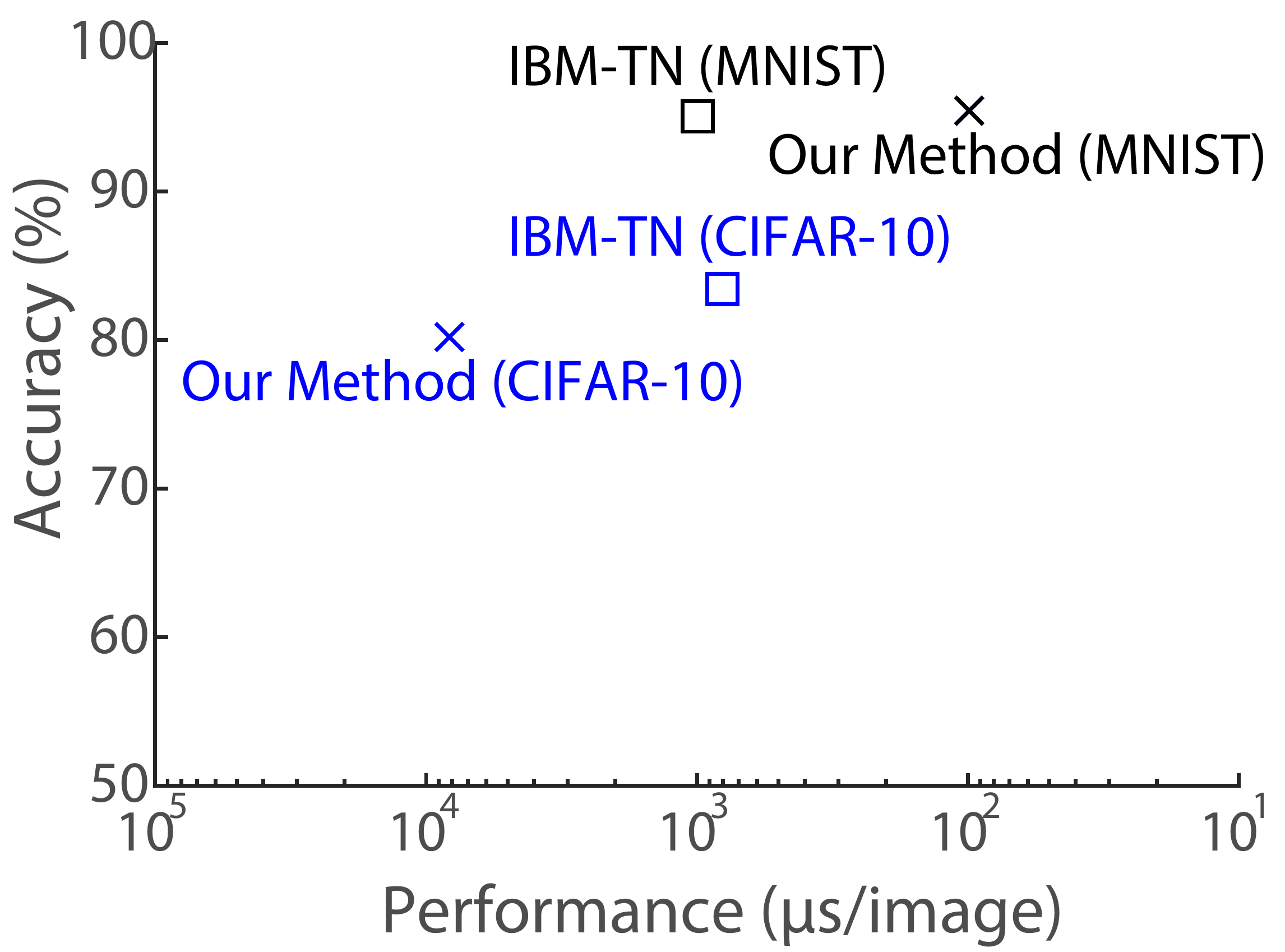}
\caption{Performance vs. accuracy results comparison
on the MNIST and CIFAR-10 benchmarks.}
\label{fig_result}
\end{center}
\vspace{-0.3em}
\end{figure}

   In this section, we provide comprehensive comparison results on MNIST, CIFAR-10, and IBM TrueNorth~\cite{esser2015backpropagation,esser2016convolutional}. Our test platform consists of one or two qual-core ARM, while the IBM TrueNorth includes 4,096 ASIC cores, which is around 500-1000 times more than our testing platform. 
%    It is impressive that the proposed FFT-based implementations
% perform as good as the state-of-the-art ASIC implementation. 
In Fig.~\ref{fig_result}, compared with IBM TrueNorth results on MNIST~\cite{esser2015backpropagation}, our model performs 10$\times$ faster than IBM TrueNorth with a little accuracy reduction on the best device result. The accuracy for IBM TrueNorth is {95\%} and the runtime is 1000\si{\us} per image on MNIST. 
Compared with IBM TrueNorth results on CIFAR-10~\cite{esser2016convolutional}, with 500-1000 times less cores, our model performs 10$\times$ slower than IMB TrueNorth.  The accuracy for IBM TrueNorth is {83.41\%} and the runtime is 800\si{\us} per image. We can see that the later work~\cite{esser2016convolutional} in 2016 on CIFAR-10 is optimized more efficiently compared with the former work~\cite{esser2015backpropagation} in 2015. Although our mobile phone based framework achieves lower performance compared with IBM TrueNorth on CIFAR-10, it is still reasonably good result considering the dramatic difference in computational resources. These results have demonstrated the effectiveness of the proposed framework.  

	\section{Conclusions}
	
	This paper presented a design optimization framework for Fast Fourier Transform-based deep neural network inference on embedded system. The proposed approach results in significant reduction in storage requirement for model parameters and improves runtime without affecting accuracy significantly. Our implementation on ARM-based embedded systems achieves runtime improvement on image classification tasks compared to IBM TrueNorth.
    
\section{Acknowledgement}
%This work is funded by the National Science Foundation Awards CCF-1733701, CNS-1704662, CNS 1739748, and CCF-1733834. 
This work is supported by the National Science Foundation funding awards CNS-1739748 and CNS-1704662.

\balance

\bibliographystyle{IEEEtran}
\bibliography{IEEEabrv,FFTDNN}

\end{document}